\pdfoutput=1
\newcommand{\ggrev}[2]{\textcolor{black}{#2}} 
\documentclass[11pt]{article}
\usepackage[table,xcdraw]{xcolor}
\usepackage{booktabs}
\usepackage{multirow}
\usepackage{graphicx}
\usepackage[]{ACL2023}
\usepackage{cuted, ragged2e}
\usepackage{times}
\usepackage{latexsym}
\usepackage{svg}
\usepackage{lscape}
\usepackage{floatrow}
\usepackage{tabularx}
\usepackage{tabto}
\usepackage{listings}
\usepackage{multirow}
\def\rot{\rotatebox}
\usepackage{caption}
\usepackage{subcaption}
\definecolor{Mycolor1}{HTML}{004488}
\definecolor{Mycolor2}{HTML}{DDAA33}
\definecolor{Mycolor3}{HTML}{BB5566}

\usepackage[T1]{fontenc}

\usepackage[utf8]{inputenc}
\usepackage{pgfplots}
\pgfplotsset{compat=1.9}

\usepackage{microtype}
\usepackage{inconsolata}
\usepackage{capt-of}
\usepackage{adjustbox}
\usepackage{caption}
\usepackage{subcaption}
\definecolor{Mycolor1}{HTML}{004488}
\definecolor{Mycolor2}{HTML}{DDAA33}
\definecolor{Mycolor3}{HTML}{BB5566}

\usepackage{todonotes}

\title{Metric-Based In-context Learning: A Case Study in Text Simplification}

\author{Subha Vadlamannati \\
  Mercer Island High School \\
  Seattle, USA\\
  \texttt{subhavee2@gmail.com} \\\And
  Gözde Gül Şahin \\
  Computer Engineering Department \\
  Koç University, Istanbul, Turkey \\ 
  \texttt{gosahin@ku.edu.tr} }

\begin{document} 
\maketitle
\begin{abstract}
In-context learning (ICL) for large language models has proven to be a powerful approach for many natural language processing tasks. However, determining the best method to select examples for ICL is nontrivial as the results can vary greatly depending on the quality, quantity, and order of examples used. In this paper, we conduct a case study on text simplification (TS) to investigate how to select the best and most robust examples for ICL. We propose \textbf{M}etric-\textbf{B}ased in-context \textbf{L}earning~(MBL) method that utilizes commonly used TS metrics such as SARI, compression ratio, and BERT-Precision for selection. Through an extensive set of experiments with various-sized GPT models on standard TS benchmarks such as TurkCorpus and ASSET, we show that examples selected by the top SARI scores perform the best on larger models such as GPT-175B, while the compression ratio generally performs better on smaller models such as GPT-13B and GPT-6.7B. Furthermore, we demonstrate that MBL is generally robust to example orderings and out-of-domain test sets, and outperforms strong baselines and state-of-the-art finetuned language models. Finally, we show that the behaviour of large GPT models can be \textit{implicitly controlled} by the chosen metric. Our research provides a new framework for selecting examples in ICL, and demonstrates its effectiveness in text simplification tasks, breaking new ground for more accurate and efficient NLG systems.
\end{abstract}

\section{Introduction}
Text simplification~(TS) is a crucial task in natural language processing, with the goal of converting complex text into simpler, easier-to-understand one. This is particularly important for individuals who struggle with comprehending complex languages, such as second language learners or individuals with cognitive impairments~\citep{stajner-2021-automatic} and disabilities like dyslexia~\citep{10.1145/2461121.2461126} and autism~\citep{autistic}. For the aforementioned reasons, NLP community has shown great interest in the topic, introducing plenty of datasets (e.g., ASSET \citep{alva-manchego-etal-2020-asset}), models, and evaluation metrics (e.g., SARI \citep{xu-etal-2016-optimizing}). 

There have been numerous approaches to TS proposed in the literature, including non-neural or rule-based methods~\citep{NassarAH19}, machine translation approaches~\citep{xu-etal-2016-optimizing}, and finetuning of large language models~\citep{sheang-saggion-2021-controllable} on downstream task data. Recently, it has been shown that large language models such as GPT-3, are capable of in-context learning~(ICL)~\cite{brown2020language}---an emerging ability to learn from in-context samples without modifying model parameters.~\footnote{We refer the readers to \url{http://ai.stanford.edu/blog/understanding-incontext/} for a summary of in-context learning inner mechanics.} Despite its strong ability, ICL still mostly falls behind the performance of finetuning techniques~\citep{surveyICL}. 


Recent studies have shown that in-context learning is highly variable to a range of factors, such as the number of examples, quality of examples, and even the order of examples~\citep{lu-etal-2022-fantastically,kategpt,surveyICL}. To address these concerns, recent literature has proposed several techniques for selecting the most relevant examples for ICL~\citep{kategpt, SorensenRRSRDKF22,Gonen22,rubin-etal-2022-learning}. The majority of them aim to \textit{retrieve} a set of samples from the validation set that resembles the test set most by either training a separate retrieval model or utilizing an existing encoder to calculate similarities between pairs of sentences. However, adopting these techniques for text-generation tasks with multiple references is nontrivial, and the need to access to the full test set to pick examples from is not desirable, and may not always be possible in real-life scenarios.   


In order to address this problem, we propose a simple yet intuitive metric-based selection technique, which we refer to as \textbf{M}etric-\textbf{B}ased in-context \textbf{L}earning~(MBL), to perform efficient and robust in-context learning with large language models for text generation tasks. Unlike previous ICL techniques, MBL only requires access to the development set and uses more informed measures rather than requiring generating sentence embeddings or training separate specified retrieval models. Furthermore, we perform an extensive set of experiments with GPT-3 models of various sizes ($175$B, $13$B, and $6.7$B)~\footnote{Throughout the paper, GPT-$175$B, GPT-$13$B, and GPT-$6.7$B will refer to the GPT-$3$ model with $175$B, $13$B, and $6.7$B parameters respectively.}, specifically focusing on their performance for TS. We investigate utilizing commonly used TS metrics (e.g., SARI, compression ratio) for example selection and answer several research questions regarding their strengths and weaknesses on a variety of datasets and models. 
Through our experiments, we show that metric-based selection can significantly improve the performance of large language models on TS. We also demonstrate that these results are generally robust to various orderings and perform well in out-of-domain settings. This paper provides the following contributions:

\begin{itemize}
  \item We provide a naive yet effective and robust approach to selecting examples for in-context learning, a.k.a., \textit{metric-based learning}~(MBL)~\footnote{We use metric-based selection and learning interchangeably.}, and show that it achieves state-of-the-art results on two well-known benchmark datasets (TurkCorpus and ASSET when the optimal metric is used (see \S\ref{ssec:rq1}))~\footnote{We find SARI score to be optimal for large models, while, Compression Ratio (CR) achieves the best scores for the smaller models.}.

   \item We demonstrate the robustness of MBL to example ordering (see \S\ref{ssec:rq2}) and to out-of-domain test sets with some exceptions (see \S\ref{ssec:rq3}), suggesting that the order of examples and the origin of the development data are not the most important factors for MBL. 
    
    \item We show that MBL improves upon important baselines (e.g., zero-shot, random selection), state-of-the art ICL selection (e.g., KATE-GPT~\cite{kategpt}) and text simplification methods~\citep{sheang-saggion-2021-controllable} (see \S\ref{ssec:rq4}).
  
    \item \textbf{Our results suggest that GPT-175B can be \textit{implicitly controlled}} via optimal metric-based learning, i.e., BERTScore Precision-based learning optimizes BLEU, while SARI-based selection optimizes SARI scores.
    
\end{itemize}
We release all generation outputs, baseline models and evaluation scripts publicly with \url{https://github.com/NLP-KU/metric-based-in-context-learning/}.

\section{Related Work}
\label{sec:rel_work}

\paragraph{Text Simplification (TS) Methods} 
Recently, LLMs have been applied to text simplification through transfer learning approaches. For instance, \citet{qiang2020BERTLS} fine-tuned a BERT model on a text simplification dataset, achieving strong results on multiple benchmarks. Similarly, \citet{sheang-saggion-2021-controllable} introduced a transfer learning approach for text simplification using the T5 model and achieved current state-of-the-art results on standard TS benchmarks. Recent work in the TS domain has a particular focus on controllable text simplification, in which different ``control tokens'' are embedded in seq2seq models to control model outputs. This is seen in both \citet{sheang-saggion-2021-controllable} and \citet{chamovitz2022cognitive}, where a large language model (BART, T5, etc.) is modified with several control tokens, like the number of words, Levenshtein similarity, and various text rewriting operations. A vast amount of earlier systems (e.g., \citep{xu-etal-2016-optimizing}) have formulated text simplification as a machine translation task and employed neural machine translation architectures. 

\paragraph{TS Evaluation} Work on the suitability of various metrics for TS has also been an active area of discussion. While the most commonly used metric in TS is currently SARI~\citep{xu-etal-2016-optimizing}, there is a concern over the metric that best correlates with human judgment. \citet{alva-manchego-etal-2021-un} conduct a detailed analysis of several commonly used metrics in the TS field, and suggest BERTScore\_Precision as a primary metric of reference-based evaluation. Following these results, we also use BERTScore\_Precision as a metric to select examples. Recent studies~\citep{sulem-etal-2018-bleu, tanprasert-kauchak-2021-flesch} analyzing the suitability of the other two common metrics, namely BLEU \citep{papineni-etal-2002-bleu} and FKGL \citep{kincaid1975derivation}, strongly advise against these metrics for TS. For these reasons, we do not select examples based on either metric.

\paragraph{Example Selection and Ordering in ICL} While large language models like GPT-$3$ perform exceptionally well on a variety of downstream tasks, selecting examples for in-context learning is nontrivial. Research on example selection is still in early stages, and a unified approach to selecting examples for downstream tasks has not yet been proposed~\citep{surveyICL}. \citet{kategpt} propose selecting the $k$ most similar examples to the test set from the training/development set via measuring cosine distance in an embedding space (e.g., encodings from RoBERTa), and achieve strong results on various tasks like table-to-text generation. On a similar line, \citet{rubin-etal-2022-learning} introduce a more sophisticated method, where the authors train a two-step retrieval model to select ICL examples. Another set of work focus on optimizing prompts via mutual information~\citep{SorensenRRSRDKF22} or perplexity~\citep{Gonen22}, that don't require labeled sets. We consider Kate-GPT~\citep{kategpt} as the closest work to ours, since both the intuition (i.e., choosing from a labeled validation set) and approach (i.e., learning-free) are similar.

\section{Metric-based In-Context Learning}
Following the line of work for retrieving the best samples from development set~\cite{kategpt,SorensenRRSRDKF22}, we introduce a simple and intuitive technique based on employing standard evaluation metrics for selecting the examples. 

\paragraph{Task Setup} Given the list of sentences $l = [c, r_1, r_2,..., r_n]$, where $c$ is the complex and $r_i$ is the simple reference sentence; our goal is to find the best $k$ pairs, $[c, r_i]$, such that the final text simplification performance on the test set is maximized. To do so, we go through each $l$ in the development set and measure the distance between each $c$ and $r_i$ according to a metric, $m$. Finally, we pick the top $k$ pairs and fill the prompt template with the samples: ``Complex sentence: \{$c$\}, Simple sentence:\{$r_i$\}''.    

We initially considered a long list of task-specific as well as general generation metrics that contain the standard evaluation metrics for TS, namely as SARI, BLEU, FKGL; as well as a simple analysis metric: Compression Ratio, and a more recent textual similarity metric BERTScore as suggested by \citet{alva-manchego-etal-2021-un}. Following the criticisms by \citet{sulem-etal-2018-bleu} and \citet{tanprasert-kauchak-2021-flesch}, we removed BLEU and FKGL from the list of candidate metrics.

\paragraph{Compression Ratio (CR)} It is simply calculated by dividing the number of characters in $c$ by the number of characters $r_i$. We consider the pairs with higher compression ratios as more preferable candidates for TS.

\paragraph{BERTScore Precision (BP)~\citep{bertscore}} BERTScore computes the cosine similarity between each token in the candidate, $y$, and reference, $x$, sentences. Precision is calculated as:
\begin{equation}
 \textnormal{Prec} = \frac{1}{|y|} \sum_{y_j \in y} \max_{x_i \in x}  x_{i}^\top y_{j}  
\end{equation}
We discard pairs with a score of 1 since they would simply be duplicates.

\paragraph{SARI~\citep{xu-etal-2016-optimizing}} It is the defacto standard evaluation metric for TS. In general terms, it compares prediction against both the input and the reference sentences. It calculates a weighted average of F1 scores for three operations: addition, deletion, and keeping. Precision and recalls for each operation are calculated based on n-gram overlaps between the prediction, input and reference sentences. To calculate the SARI score for each $c$-$r_i$ pair, we denote $r_i$ as the prediction, $c$ as the input, and $[r_1,...,r_{i-1}, r_{i+1},..., r_n]$ as the reference sentences. Hence, this measure can only be applied when there are multiple references.

\section{Experimental Setup}
\label{sec:setup}
To investigate the effects of metric-based selection techniques on TS, we perform a comprehensive set of experiments using various LLMs, sample sizes, and datasets; and compare against strong baseline and state-of-the-art models. Following the criticism~\cite{sulem-etal-2018-bleu} on using BLEU~\citep{papineni-etal-2002-bleu}, we use SARI~\citep{xu-etal-2016-optimizing} as our main evaluation metric. However, we also report BLEU for two reasons: i) to be consistent with previous works~(see~\S \ref{ssec:sota}) and ii) to gain more insights on how the chosen metric for MBL effects the results measured with different metrics. 

\subsection{Models}
Due to its recent success in text generation and in-context learning for various downstream tasks, we experiment with the GPT-3~\citep{gpt3} model. We use three different version with the following parameter sizes: 175B, a.k.a., \texttt{da-vinci-003}, 13B, a.k.a., \texttt{curie}, and 6.7B, a.k.a., \texttt{babbage}. 
We used OpenAI API\footnote{https://openai.com/} to generate responses using temperature=$0.7$, max\_tokens=$256$ top\_p=$1$, frequency penalty=$0$ and presence penalty = $0$.

\subsection{Datasets}
    We perform our main experiments on the ASSET~\cite{alva-manchego-etal-2020-asset} and TurkCorpus~\cite{xu-etal-2016-optimizing} datasets. To investigate the transferability of our models, we conduct additional experiments on an out-of-domain cognitive simplification dataset, FestAbility~\cite{chamovitz2022cognitive}.
    
    \paragraph{TurkCorpus} is a widely-used dataset with 2000 validation and 359 test sentences. It has 8 reference sentences for each original sentence in both the validation and test set. 
     
    \paragraph{ASSET} is another widely used TS dataset with the intention of improving upon TurkCorpus. It has the same 2000 validation and 359 original test sentences but introduces 10 new reference sentences for each original sentence. ASSET is deemed to be simpler by human evaluation in both fluency and simplicity ~\citep{alva-manchego-etal-2020-asset}. ASSET improves upon TurkCorpus as it allowed human reviewers to focus on a wider variety of TS operations, which are: lexical paraphrasing, compression, and sentence splitting. Because of this, we emphasize the experiments done on ASSET rather than TurkCorpus while interpreting the results and answering the research questions in \S \ref{sec:setup}. 
    
    \paragraph{FestAbility} is a cognitive simplification dataset with 321 pairs of complex and simple sentences---i.e., only one reference sentence. Each of these is additionally annotated with rewriting operations such as \texttt{<ADDITION>} and \texttt{<DELETION>}. These sentences are generated from the transcript of the virtual accessibility conference, and simplifications are generated from the Yalon Method~\citep{chamovitz2022cognitive}, a specialized method for simplifying text for individuals with cognitive impairments.

\subsection{Baselines}

    For comparison, we implement three baselines: i) random selection ii) KATE-GPT~\citep{kategpt} and iii) zero-shot. In the random setting, we randomly select $c$ and $r_i$ pairs from the validation sets. For KATE-GPT~\citep{kategpt}, we use the default setting that employs RoBERTa-base for contextualized embeddings and cosine similarity for the distance metric. Given that KATE-GPT calculates sentence pair similarities between the development and test set, unlike just the development set (like ours), we choose complex sentences as the representative. Zero-shot setting is simply conducted with the same instruction prompt without providing any examples.

\subsection{Text Simplification State-of-the-art}
\label{ssec:sota}
    We compare our results across multiple state-of-the-art systems.
    
    \paragraph{MUSS (BART+ACCESS Supervised)} \citet{martin-etal-2022-muss} fine-tune BART~\citep{bart} and add information from the four simplification tokens trained in ACCESS. 
    
    \paragraph{Finetuned-T5} \citet{sheang-saggion-2021-controllable} finetune T5 by adding multiple control tokens (e.g., compression ratio, Levenshtein similarity ratio, word rank, and number of words) similar to ACCESS, which control the model's outputs. To the best of our knowledge, they achieve the current state-of-the-art on both the TurkCorpus and ASSET datasets, with SARI scores of 43.31 and 45.04 respectively.

\subsection{Evaluation}
Even though SARI is considered the standard evaluation metric in our experiments below, we evaluate the results both with SARI and BLEU to emphasize the behavior differences in metric-based selection. It should be noted that, SARI compares prediction against input and reference(s), while BLEU compares only the prediction against reference(s). We use the package EASSE~\citep{alva-manchego-etal-2019-easse} with the default settings~\footnote{We used the BLEU with $n=5$ against all references provided. The implementation is available at \url{https://github.com/feralvam/easse}.} to generate reports for all of our experiments.

\section{Experiments and Results}
\begin{figure}[]
    \centering
    \includegraphics[width=\linewidth]{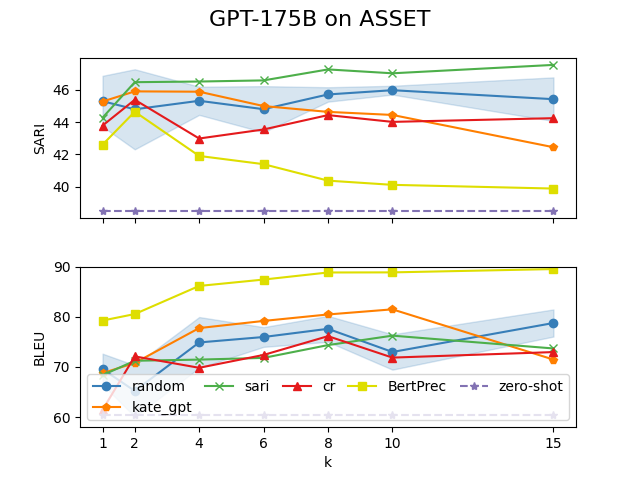}
    \caption{GPT-$175$B results on ASSET. Top: SARI scores, Bottom: BLEU scores.}
    \label{fig:davinci_asset_all}
\end{figure}
We conduct a comprehensive set of experiments with the setup explained in \S \ref{sec:setup}. Following the work by \citet{lu-etal-2022-fantastically}, we experiment with the $k$ values as $1,2,4,6,8,10,15,20$ examples. We repeat the random baseline experiments three times for each $k$. We aim to answer the following research questions (RQ):

\paragraph{RQ1:} How do different metric-based selection techniques compare?~(\S\ref{ssec:rq1})

\paragraph{RQ2:} Is metric-based sample selection robust to the order of the prompts?~(\S\ref{ssec:rq2}) 

\paragraph{RQ3:} How does metric-based ICL compare to state-of-the-art text simplification methods?~(\S\ref{ssec:rq3})

\paragraph{RQ4:} Does metric-based selection performance on one dataset transfer to other out-of-domain datasets?~(\S\ref{ssec:rq4})

\subsection{RQ1: Effect of Metrics} 
\label{ssec:rq1}
Our main results with our default settings (GPT-$175$B on ASSET) is shown in Fig.~\ref{fig:davinci_asset_all}. First of all, we observe that the random baseline is quite strong on average, however, with a \textbf{large variation} for most $k$ values; while zero-shot results are quite weak for all datasets. Interestingly, SARI-based selection consistently leads to the highest SARI scores for $k>1$, while BERTPrec-based selection gives the highest BLEU and lowest SARI scores consistently for each $k$. Kate-GPT follows BERTPrec-based selection for the BLEU score, while providing results on par or lower than the random baseline for the SARI score. 

\begin{figure}[]
    \centering
    \includegraphics[width=\linewidth]{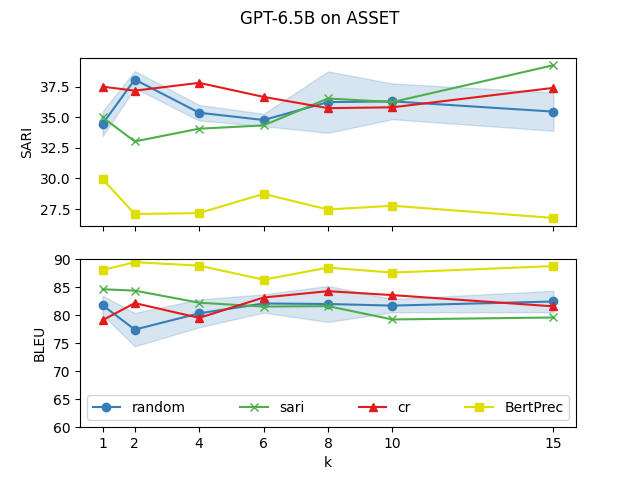}
    \caption{GPT-6.7B results on ASSET. Top: SARI scores, Bottom: BLEU scores.}
    \label{fig:babbage_asset_all}
\end{figure}

Next, we check whether our findings hold for smaller models. In Fig.~\ref{fig:babbage_asset_all}, we show the results of our smallest model, GPT-6.7B on the ASSET dataset. Since the zero-shot results were significantly lower than $k=1$, we show them in Table~\ref{tab:zeroShot}, rather than plotting. Not surprisingly the highest SARI scores are achieved via the largest model; however, the opposite is not true for BLEU. The smallest model achieves the highest BLEU scores that raises another warning flag for using BLEU for TS evaluation. 

Similar to the larger model, BERTPrec-based selection achieves the highest BLEU, and the lowest SARI scores. SARI-based selection provides considerably high SARI scores only for larger $k$s, suggesting the implicit controlling mechanism does not exist, or is only triggered with more samples. We also observe that CR performs relatively better on GPT-6.7B which suggests compression provides a stronger signal (e.g., deletion, shorter tokens) that can be utilized better by smaller models for simplification. 

\begin{figure*}[ht]
    \centering  
    \includegraphics[scale=0.45]{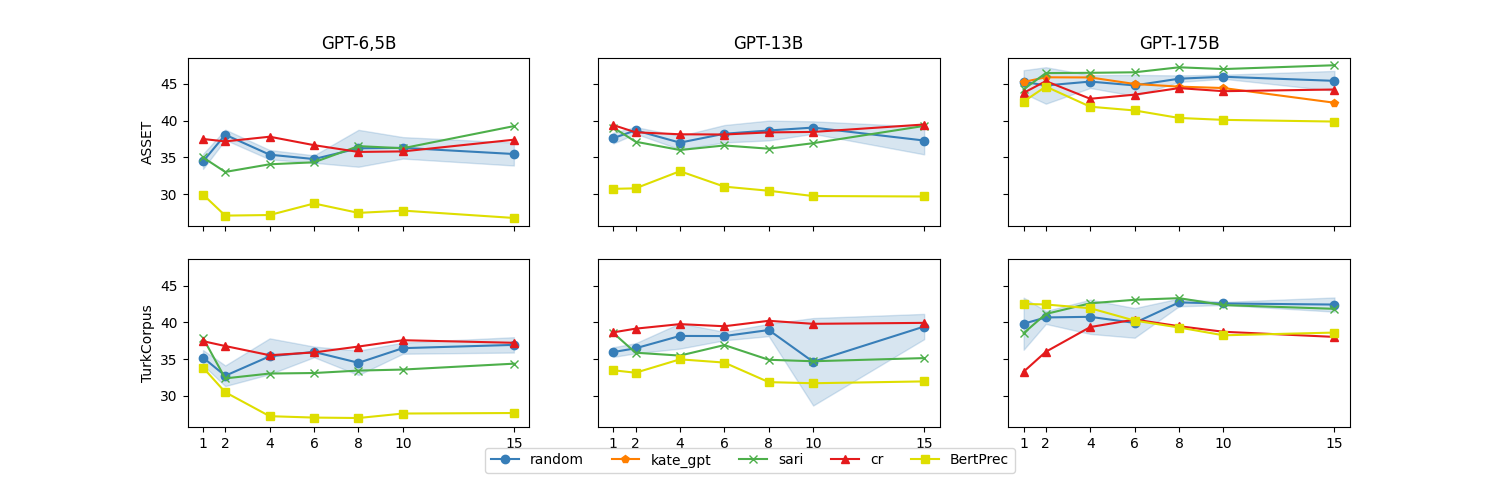}
    \caption{SARI scores for GPT-$6.7$B, GPT-$13$B and GPT-$175$B models on ASSET (top) and TurkCorpus (bottom) datasets. See App.~\ref{sec:bleuresults} for BLEU scores.}
    \label{fig:dataset_all_models}
\end{figure*}

Finally, we investigate how the quality of the dataset affects the metric-based selection techniques, i.e., whether they are robust to noise. Fig~\ref{fig:dataset_all_models} shows an overview of the SARI scores from all models on the noisy (i.e., TurkCorpus) and the cleaner (i.e., ASSET) dataset. Even though the general patterns are visible, the results on TurkCorpus are moderately less conclusive. 

\begin{table}[!htp]
\begin{center}
\scalebox{0.62}{
\begin{tabular}{lccrr} 
\toprule
\textbf{Dataset}& \textbf{Model} & \textbf{SARI} & \textbf{BLEU} \\
\midrule
\multirow{3}{*}{TurkCorpus} 
& GPT-$175$B & \textbf{32.17} & 42.34 \\
& GPT-$17$B & 27.19 & 38.35 \\
& GPT-$6.7$B & 24.13 & \textbf{57.14} \\
\midrule
\multirow{3}{*}{ASSET} 
& GPT-$175$B & \textbf{38.49} & 60.48 \\
& GPT-$17$B & 30.45 & 40.13 \\
& GPT-$6.7$B & 26.28 & \textbf{69.83} \\
\bottomrule
\end{tabular}
\caption{Zero-shot results} 
\label{tab:zeroShot}
}
\end{center}
\end{table}

\paragraph{Selection Metric versus Evaluation Metric} Even though this was not one of our main research questions, we observe a strong relation between the metric used for MBL and the metric used for evaluation. For all the model and dataset size settings, we observe that BLEU scores are consistently higher when the examples are selected via BERTScore\_Precision. When we evaluate with the SARI score, SARI-metric behaves similarly for the GPT-175B model, however CR-metric performs better for the smaller models. More evidence for the relation between BLEU and BERTScore\_Precision can be found in Appendix~\ref{sec:bleuresults}. This suggests that the behaviour of large-enough GPT models can be \textit{implicitly controlled} via MBL, which paves the way to a new research direction and needs further investigation. 

\subsection{RQ2: Effect of Order}
\label{ssec:rq2}

Previous research~\citep{lu-etal-2022-fantastically} has shown that the order of the examples may have a significant impact on ICL performance. \ggrev{N/A}{Commonly used orderings~\citep{lu-etal-2022-fantastically} include sorting from highest to lowest quality example, vice versa, and random selection.} Inspired by these findings, we investigate the robustness of our selection metrics across sample orders. To do so, for each metric we perform three different order arrangements, namely as highest $\rightarrow$ lowest, lowest $\rightarrow$ highest, and random ordering for each metric. To have enough variation, we only experiment with $k=6,8,10,15$. As the baseline, we randomly pick samples and arrange them in 3 different randomized orders. 
\begin{figure*}[]
    \centering
    \begin{subfigure}[b]{0.47\textwidth} \includegraphics[width=\linewidth]{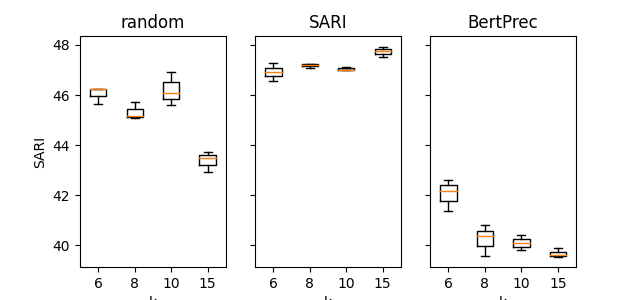}
    \end{subfigure}
    \begin{subfigure}[b]{0.47\textwidth}
       \includegraphics[width=\linewidth]{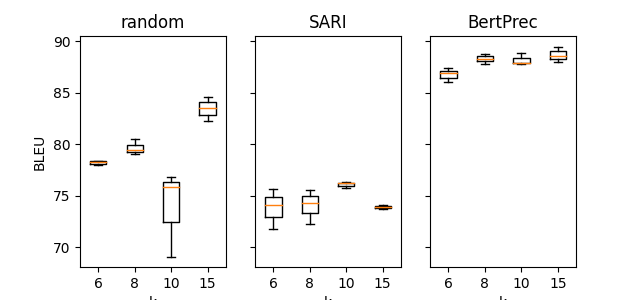}
    \end{subfigure}
    \caption{Boxplots for GPT-$175$B model performance on ASSET with sample (re)ordering via random, SARI-based and BERTPrec-based selections. Performance shown in SARI (left), and BLEU (right) }
    \label{fig:boxplot_davinci_asset}
\end{figure*}

In Fig~\ref{fig:boxplot_davinci_asset}, we show how the performance of GPT-$175$B varies on ASSET when the samples that are i) picked randomly, ii) by SARI-based selection and iii) by BERTPrec-based selection are reordered following the above setup. As can be seen, the best-performing metrics, are also the most robust compared to others. To elaborate, SARI-based selection that provided the highest SARI scores has the least variation, i.e., most robust to order; while BERTPrec-based selection provides the most stable BLEU scores along with the highest.  


\subsection{RQ3: Comparison to State-of-the-art}
\label{ssec:rq3}
Finally, we compare our best and average model settings to state-of-the-art fine-tuned models~\footnote{The models which do not provide SOTA (e.g., KATE) are not included in the Table. The statistical significance cannot be provided since there is only one setting for the few-shot setting.}. The results are given in Table~\ref{tab:sota}. Here, Random and SARI averages are calculated from \S\ref{ssec:rq1} results, averaged over all $k$, with random selection being additionally averaged over all three random selections. These averages are reported for GPT-$175$B results, because it is generally the best model when considering averages across both datasets. As can be seen, the GPT-$175$B model with SARI-based selection outperforms existing results on all datasets, followed by random best and SARI-averaged. The exact settings (number of examples, model, and ordering) for SARI and Random Best can be found Appendix~\ref{sec:optimalsettings}.

\begin{table}[]
\begin{center}
\scalebox{0.6}{
\begin{tabular}{p{0.7\linewidth}ccc} 
\toprule
\textbf{Model}  & \textbf{ASSET} & \textbf{TurkCorpus} & \textbf{FestAbility} \\
\midrule
Finetuned T5                    & 45.04          & 43.31               & N/A         \\ 

MUSS (BART + ACCESS) & 43.63          & 42.62               & N/A         \\

BART-Large+Classifier           & 38.76          & N/A                 & 27.13       \\

\midrule
(Ours) Random-Best 
& 46.93          & 43.14               & N/A         \\

(Ours) Random-Average                  & 45.33          & 40.32               & N/A         \\
\midrule

(Ours) MBL-Best  
& \textbf{47.94} & \textbf{43.46}      & \textbf{44.86} \\

(Ours) MBL-Average                    & 46.63          & 41.78               & 43.55      \\
\bottomrule
\end{tabular}
\caption{Comparison to TS state-of-the-art models. Random- and MBL-best examples are selected from the top examples in all experiments run. 
The best results are shown in bold. For more information on the exact settings for MBL and Random Best, see Appendix~\ref{sec:optimalsettings}.} 
\label{tab:sota}
}
\end{center}
\end{table}

\subsection{RQ4: Task Transfer}
\label{ssec:rq4}

In order to evaluate the suitability of our approach for unseen tasks and datasets, we experiment with choosing samples from a tune set and testing the performance on an unseen set. Here, we use ASSET and TurkCorpus as the tune set and evaluate on all three datasets: ASSET, TurkCorpus, and FestAbility. To investigate different metrics and language models, we perform experiments with GPT-$175$B with SARI-based selection and GPT-$13$B with CR-based selection as both metric selection techniques are generally best on those respective models. We use the best experimental settings from Table \ref{tab:sota} in out-of-domain settings, comparing them with their in-domain counterparts. For example, the best setting for TurkCorpus is k=$6$ with high to low ordering (see Appendix \ref{sec:optimalsettings} for more details on optimal experiment settings), so we compare the results of the model when given this setting on both the TurkCorpus and ASSET datasets. State-of-the-art results in this table refer to the best setting for in-domain experiments (i.e. ASSET evaluated on ASSET or TurkCorpus evaluated on TurkCorpus).  

Our results are given in Table~\ref{tab:ICL}. For easy comparison, the Table also includes in-domain selection results as well as the current state-of-the-art scores taken from Table~\ref{tab:sota}. In the first row, we observe that samples selected from TurkCorpus and tested on ASSET achieve significantly lower SARI scores than their in-domain variant for the GPT-$175$B setting, whereas the gap is lower for the GPT-$13$B. On the other hand, for the TurkCorpus test setting (second row), we see that GPT-$175$B model prompted with the best ASSET examples achieves even better results than the in-domain setting, suggesting a highly successful transfer. This ability cannot be observed for the GPT-$13$B model with CR-based selection. The final row shows the transfer results to another related but different task. It is apparent that both models prompted with ASSET examples achieve marginally higher scores than the TurkCorpus ones.

Taking a look at the BLEU scores, we see that out-of-domain configurations on the TurkCorpus and ASSET datasets generally tend to match or even exceed their in-domain counterparts, suggesting a successful transfer. However, on the FestAbility dataset, we observe notably low BLEU scores, which are in-part due to the nature of FestAbility, in which sentences are often simplified in unconventional ways. Additionally, FestAbility sentences are extremely short, with only 1452 unique tokens in the original sentences and 996 unique tokens in the simplified sentences \citep{chamovitz2022cognitive}, leading to unconventional results.

\begin{table}[!htp]
\begin{center}
\scalebox{0.62}{
\begin{tabular}{lccrr} 
\toprule
\textbf{Test Set}& \textbf{Model Setting} & \textbf{Tune Set} & \textbf{SARI} & \textbf{BLEU} \\
\midrule
\multirow{4}{*}{\rot{90}{ASSET}} & GPT-$175$B, SARI, high to low, k=6 & TurkCorpus & 43.46 & 79.83 \\
& GPT-$175$B, SARI, high to low, k=6 & ASSET & 46.93 & 75.67 \\
& GPT-$13$B, CR, high to low, k=15 & TurkCorpus & 41.73 & 74.57 \\
& GPT-$13$B, CR, high to low, k=15 & ASSET & 41.9 & 76.49 \\
& \textit{State-of-the-art (MBL-Best)} &  & \textit{47.94} & \textit{73.92} \\
& \textit{Zero-shot (GPT-$175$B)} &  & \textit{38.49} & \textit{60.48} \\
\midrule
\multirow{4}{*}{\rot{90}{TurkCorpus}} & GPT-$175$B, SARI, random, k=15 & ASSET & 42.37 & 64.52 \\
& GPT-$175$B, SARI, random, k=15 & TurkCorpus & 41.48 & 85.89 \\
& GPT-$13$B, CR, high to low, k=15 & ASSET & 39.44 & 71.15 \\
& GPT-$13$B, CR, high to low, k=15 & TurkCorpus & 40.37 & 73.83 \\
& \textit{State-of-the-art (MBL-Best)} &  & \textit{43.46} & \textit{79.83} \\
& \textit{Zero-Shot (GPT-$175$B)} &  & \textit{32.17} & \textit{42.34} \\
\midrule
\multirow{5}{*}{\rot{90}{FestAbility}} & GPT-$175$B, SARI, random, k=6 & TurkCorpus & 42.24 & 20.76 \\
& GPT-$175$B, SARI, random, k=15 & ASSET & 44.86 & 17.08 \\
& GPT-$13$B, CR, high to low, k=15 & TurkCorpus & 25.46 & 23.37 \\
& GPT-$13$B, CR, high to low, k=15 & ASSET & 36.63 & 12.01 \\
& \textit{State-of-the-art (MBL-Best)} &  & \textit{44.86} & N/A \\
& \textit{Zero-shot (GPT-$175$B)} &  & \textit{40.77} & \textit{6.9} \\
\bottomrule
\end{tabular}
\caption{ICL out-of-domain results for GPT-$175$B, SARI-based selection and GPT-$13$B, CR-based selection. Examples are picked from the \textit{Tune Set} and tested on the \textit{Test Set}. Zero-shot results are from GPT-$175B$ and given in the final rows for each dataset.} 
\label{tab:ICL}
}
\end{center}
\end{table}

\section{Qualitative Analysis}
\label{sec:discussion}
\definecolor{Mycolor1}{HTML}{004488}
\definecolor{Mycolor2}{HTML}{DDAA33}
\definecolor{Mycolor3}{HTML}{BB5566}

\begin{table*}[htbp]
         \centering
            \small
             \renewcommand{\arraystretch}{1.5}
        \begin{tabular}{p{0.2\linewidth} p{0.75\linewidth}} \toprule
        Metric   & Top 2 Examples \\   \midrule                                                                                                     
        Compression Ratio   & \textbf{Complex Sentence} They \textcolor{Mycolor1}{manifest} with either \textcolor{Mycolor2}{neurological complications} or with skin problems (or \textcolor{Mycolor3}{occasionally} both).  \newline                                   \textbf{Simple Sentence}  They \textcolor{Mycolor1}{show} either \textcolor{Mycolor2}{brain} or skin \textcolor{Mycolor2}{problems} (or both).                                                                            \\ \hline
                            & \textbf{Complex Sentence} The \textcolor{Mycolor1}{psychological state of sympathy} is closely linked with \textcolor{Mycolor3}{that of} compassion, empathy and \textcolor{Mycolor3}{empathic concern}. \newline \textbf{Simple Sentence}                             \textcolor{Mycolor1}{Sympathy} is closely linked with compassion and empathy.                                                                       \\
        \midrule
        BertScore Precision & \textbf{Complex Sentence} Sthenurine forelimbs were long with two extra-long fingers and claws compared with the \textcolor{Mycolor3}{relatively} small, stiff arms of modern macropods.  \newline \textbf{Simple Sentence}  Sthenurine forelimbs were long with two extra-long fingers and claws compared with the small, stiff arms of modern macropods. \\ \hline
                            & \textbf{Complex Sentence} In 1828, Coenraad Johannes van Houten \textcolor{Mycolor1}{developed} the first cocoa powder producing machine in the Netherlands.    \newline \textbf{Simple Sentence}                            In 1828, Coenraad Johannes van Houten \textcolor{Mycolor1}{created} the first cocoa powder producing machine in the Netherlands.                    \\
        \midrule
        SARI                & \textbf{Complex Sentence} The organic matter in soil \textcolor{Mycolor1}{derives} from plants and animals.  \newline \textbf{Simple Sentence}                                                                               The organic matter in soil \textcolor{Mycolor1}{comes} from plants and animals.                                                                     \\ \hline
                            & \textbf{Complex Sentence} Dennis Lee Hopper (born May 17, 1936) is an American actor, filmmaker and artist.                                           \newline       \textbf{Simple Sentence}        Dennis Lee Hopper was born on May 17, 1936. \underline{He is an American actor, filmmaker and artist.}     \\
        \bottomrule
        \end{tabular}
        \caption{Top 2 examples from each applicable selection metric (random and KATE-GPT selection were not applicable). All samples taken from the ASSET Validation dataset. We color rephrases first in \textcolor{Mycolor1}{blue} and then in \textcolor{Mycolor2}{yellow}, mark significant deletions in \textcolor{Mycolor3}{red}, and \underline{underline} sentence splits.} 
     \label{tab:top2examples}
    \end{table*}


In this section, we perform a qualitative analysis of different model generated simplifications and metric-based prompting examples in order to better understand how different settings affect model outputs. 

\subsection{Explaining Performance as $k$ Increases}
We aim to understand why certain metrics (BERTPrec and KATE-GPT) tend to perform worse as $k$ increases, while other metrics (SARI) tend to perform better as $k$ increases when evaluated on SARI scores (as seen in Fig.~\ref{fig:dataset_all_models}). \ggrev{old text}{In fact, this result is commonly seen in other papers~\citep{zhao2021calibrate,ZhangFT22}, where they describe that adding more training examples can sometimes hurt accuracy.} By analyzing output of metric-based selection on a fixed dataset and model (ASSET, GPT-$175$B) seen in Appendix~\ref{ssec:dv3asset}, we aim to understand the performance of different metrics as the number of examples, or $k$, increases. Our analysis focuses on three different metrics (KATE-GPT, BERTPrec, and SARI) and a particularly difficult example due to its unconventional subject nature, multiple abbreviations, and unknown words, and objectively confusing sentence structure. In general, we see from earlier trends that KATE-GPT and BERTPrec-selected examples tend to get worse (w.r.t SARI) as $k$ increases (see Figure~\ref{fig:davinci_asset_all}-top). We also observe this qualitatively, as $k$ increases, KATE-GPT and BERTPrec examples become closer to the original sentence, with BERTPrec generations even matching the original sentence at $k=15$. 
However, as the value of k increases, SARI-selected examples show an improvement in quality. We observe that examples selected using the SARI score metric tend to: i) split sentences more frequently, and ii) decode potentially confusing abbreviations, such as ``OEL''.

\textbf{Sentence Splitting: } SARI-selected examples are more prone to splitting sentences (see $k$=2,15), which may be in-part due to the style of the top SARI examples, which include sentence splitting; while this is not present in any of the other metrics. See Appendix \ref{ssec:dv3asset} for examples. Sentence splitting is correlated with increased human comprehension of TS outputs~\cite{williams-etal-2003-experiments}. This is particularly interesting because it leads us to infer that models can potentially learn the ``style'' of the reference sentences. 

\textbf{Abbreviations:} In all three cases, ($k=2,8,15$) examples selected by SARI score remove the potentially confusing abbreviation ``OEL'' and instead replace it with either ``original English-language'' or ``English-language'', while KATE-GPT and BERTPrec selected examples only exhibit this behavior for $k=2$ (see Appendix~\ref{ssec:dv3asset}).

\subsection{Analyzing Model Size}
Model size plays a significant role in output sentences, with smaller models (especially GPT-$6.7$B) tending to change very little structurally from the original sentence, regardless of the metric used to select examples. See Appendix \ref{sec:modeloutputs} for a complete list of model outputs on all metrics for the original sentence ``OEL manga series Graystripe's Trilogy There is a three volume original English-language manga series following Graystripe, between the time that he was taken by Twolegs in Dawn until he returned to ThunderClan in The Sight''. From these results, we conclude that GPT-$6.7$B tends to hardly change sentences at all, with both Random and BERTPrec-selected examples having no change from the original sentence. SARI-selected examples adds a comma, but CR-selection prompts the model to rephrase key parts of the sentence. GPT-$13$B performs considerably better when looking at a qualitative analysis, as all examples have removed ``OEL manga series Graystripe's Trilogy'' and restructured the sentence to be more concise, and SARI-selected going as far to remove an ambiguous abbreviation ``OEL''. These qualitative observations are consistent with our results from Figure~\ref{fig:davinci_asset_all}.

\subsection{Analyzing Top Metric-Selected Examples}
In this section, we analyze the top metric-selected examples for compression ratio, BERTPrec, and SARI. In Table~\ref{tab:top2examples} we include the top $2$ examples for each metric from the ASSET validation dataset, and in Appendix~\ref{sec:topasset} we include the remaining top $8$ examples for SARI and BERTPrec selection. 

Looking at the style of both BERTPrec and SARI score, both metrics' top examples barely change from the original sentences, often only changing one or two words (i.e., movie $\rightarrow$ film) but leaving the rest unchanged, primarily using deletion or rewriting operations. However, in CR top examples, we see extreme deletions from the original sentences and several rewriting operations done (which is consistent with our understanding of the compression ratio). Notably, we also see that top SARI examples are the only metric that use sentence splitting (see the $2$nd example under SARI from Table~\ref{tab:top2examples}).

\section{Conclusion and Future Work}
In conclusion, we propose a novel and robust method for selecting examples in the TS domain, evaluating its effectiveness on multiple well-known TS datasets and even on downstream tasks like cognitive simplification. Our experiments demonstrate state-of-the-art results in the field of TS and CS, reaching scores of 44.86 on FestAbility, 47.94 on ASSET and 43.46 on TurkCorpus. We hope that future work will generalize our findings in other text generation tasks and other domains. 

\section*{Limitations}

Our approach is computationally and financially intensive, especially on the GPT-175B model, which limits its scalability to smaller, open-source models. While our approach has shown strong results in the TS domain, we are not yet sure whether using domain-specific selection methods is widely applicable. Our approach also is not applicable in true few-shot settings in which a large validation set is not available to select examples from. Also, our approaches' scalability to other downstream TS tasks outside of cognitive simplification is yet to be tested, especially in different domains. We tested on two well-known TS datasets (ASSET and TurkCorpus), and we did not test on another known TS dataset, Newsela, due to its restrictive licensing. Additionally, we have tested our approach on example numbers up to $15$ due to financial constraints, and testing on higher numbers of examples may show additional insights and we leave this for future researchers.

\section*{Ethics Statement}
We acknowledge that while our approach reaches high scores on datasets aimed for individuals with disabilities, further research and evaluation from humans with specific disabilities listed in this paper is crucial to determine the true effectiveness of our approach in these scenarios.

\section*{Acknowledgements}
This work has been supported by the Scientific and Technological Research Council of Türkiye~(TÜBİTAK) as part of the project ``Automatic Learning of Procedural Language from Natural Language Instructions for Intelligent Assistance'' with the number 121C132. We also gratefully acknowledge the Fatima Fellowship and KUIS AI Lab for providing support. We thank our anonymous reviewers and the members of GGLab who helped us improve this paper.

\bibliography{custom}

\begin{thebibliography}{29}
\expandafter\ifx\csname natexlab\endcsname\relax\def\natexlab#1{#1}\fi

\bibitem[{Alva-Manchego et~al.(2020)Alva-Manchego, Martin, Bordes, Scarton,
  Sagot, and Specia}]{alva-manchego-etal-2020-asset}
Fernando Alva-Manchego, Louis Martin, Antoine Bordes, Carolina Scarton,
  Beno{\^\i}t Sagot, and Lucia Specia. 2020.
\newblock \href {https://doi.org/10.18653/v1/2020.acl-main.424} {{ASSET}: {A}
  dataset for tuning and evaluation of sentence simplification models with
  multiple rewriting transformations}.
\newblock In \emph{Proceedings of the 58th Annual Meeting of the Association
  for Computational Linguistics}, pages 4668--4679, Online. Association for
  Computational Linguistics.

\bibitem[{Alva-Manchego et~al.(2019)Alva-Manchego, Martin, Scarton, and
  Specia}]{alva-manchego-etal-2019-easse}
Fernando Alva-Manchego, Louis Martin, Carolina Scarton, and Lucia Specia. 2019.
\newblock \href {https://doi.org/10.18653/v1/D19-3009} {{EASSE}: Easier
  automatic sentence simplification evaluation}.
\newblock In \emph{Proceedings of the 2019 Conference on Empirical Methods in
  Natural Language Processing and the 9th International Joint Conference on
  Natural Language Processing (EMNLP-IJCNLP): System Demonstrations}, pages
  49--54, Hong Kong, China. Association for Computational Linguistics.

\bibitem[{Alva-Manchego et~al.(2021)Alva-Manchego, Scarton, and
  Specia}]{alva-manchego-etal-2021-un}
Fernando Alva-Manchego, Carolina Scarton, and Lucia Specia. 2021.
\newblock \href {https://doi.org/10.1162/coli_a_00418} {The (un)suitability of
  automatic evaluation metrics for text simplification}.
\newblock \emph{Computational Linguistics}, 47(4):861--889.

\bibitem[{Barbu et~al.(2015)Barbu, Martín-Valdivia, Martínez-Cámara, and
  López}]{autistic}
Eduard Barbu, Maria Martín-Valdivia, Eugenio Martínez-Cámara, and L.~López.
  2015.
\newblock \href {https://doi.org/10.1016/j.eswa.2015.02.044} {Language
  technologies applied to document simplification for helping autistic people}.
\newblock \emph{Expert Systems with Applications}, 42:5076–5086.

\bibitem[{Brown et~al.(2020{\natexlab{a}})Brown, Mann, Ryder, Subbiah, Kaplan,
  Dhariwal, Neelakantan, Shyam, Sastry, Askell et~al.}]{brown2020language}
Tom Brown, Benjamin Mann, Nick Ryder, Melanie Subbiah, Jared~D Kaplan, Prafulla
  Dhariwal, Arvind Neelakantan, Pranav Shyam, Girish Sastry, Amanda Askell,
  et~al. 2020{\natexlab{a}}.
\newblock Language models are few-shot learners.
\newblock \emph{Advances in neural information processing systems},
  33:1877--1901.

\bibitem[{Brown et~al.(2020{\natexlab{b}})Brown, Mann, Ryder, Subbiah, Kaplan,
  Dhariwal, Neelakantan, Shyam, Sastry, Askell, Agarwal, Herbert-Voss, Krueger,
  Henighan, Child, Ramesh, Ziegler, Wu, Winter, Hesse, Chen, Sigler, Litwin,
  Gray, Chess, Clark, Berner, McCandlish, Radford, Sutskever, and
  Amodei}]{gpt3}
Tom~B. Brown, Benjamin Mann, Nick Ryder, Melanie Subbiah, Jared Kaplan,
  Prafulla Dhariwal, Arvind Neelakantan, Pranav Shyam, Girish Sastry, Amanda
  Askell, Sandhini Agarwal, Ariel Herbert-Voss, Gretchen Krueger, Tom Henighan,
  Rewon Child, Aditya Ramesh, Daniel~M. Ziegler, Jeffrey Wu, Clemens Winter,
  Christopher Hesse, Mark Chen, Eric Sigler, Mateusz Litwin, Scott Gray,
  Benjamin Chess, Jack Clark, Christopher Berner, Sam McCandlish, Alec Radford,
  Ilya Sutskever, and Dario Amodei. 2020{\natexlab{b}}.
\newblock \href {https://doi.org/10.48550/ARXIV.2005.14165} {Language models
  are few-shot learners}.

\bibitem[{Chamovitz and Abend(2022)}]{chamovitz2022cognitive}
Eytan Chamovitz and Omri Abend. 2022.
\newblock Cognitive simplification operations improve text simplification.
\newblock In \emph{Proceedings of the 26th Conference on Computational Natural
  Language Learning (CoNLL)}, pages 241--265.

\bibitem[{Dong et~al.(2023)Dong, Li, Dai, Zheng, Wu, Chang, Sun, Xu, Li, and
  Sui}]{surveyICL}
Qingxiu Dong, Lei Li, Damai Dai, Ce~Zheng, Zhiyong Wu, Baobao Chang, Xu~Sun,
  Jingjing Xu, Lei Li, and Zhifang Sui. 2023.
\newblock \href {https://doi.org/10.48550/ARXIV.2301.00234} {A survey for
  in-context learning}.

\bibitem[{Gonen et~al.(2022)Gonen, Iyer, Blevins, Smith, and
  Zettlemoyer}]{Gonen22}
Hila Gonen, Srini Iyer, Terra Blevins, Noah~A. Smith, and Luke Zettlemoyer.
  2022.
\newblock \href {https://doi.org/10.48550/arXiv.2212.04037} {Demystifying
  prompts in language models via perplexity estimation}.
\newblock \emph{CoRR}, abs/2212.04037.

\bibitem[{Kincaid et~al.(1975)Kincaid, Fishburne~Jr, Rogers, and
  Chissom}]{kincaid1975derivation}
J~Peter Kincaid, Robert~P Fishburne~Jr, Richard~L Rogers, and Brad~S Chissom.
  1975.
\newblock Derivation of new readability formulas (automated readability index,
  fog count and flesch reading ease formula) for navy enlisted personnel.
\newblock Technical report, Naval Technical Training Command Millington TN
  Research Branch.

\bibitem[{Lewis et~al.(2020)Lewis, Liu, Goyal, Ghazvininejad, Mohamed, Levy,
  Stoyanov, and Zettlemoyer}]{bart}
Mike Lewis, Yinhan Liu, Naman Goyal, Marjan Ghazvininejad, Abdelrahman Mohamed,
  Omer Levy, Veselin Stoyanov, and Luke Zettlemoyer. 2020.
\newblock \href {https://doi.org/10.18653/v1/2020.acl-main.703} {{BART:}
  denoising sequence-to-sequence pre-training for natural language generation,
  translation, and comprehension}.
\newblock In \emph{Proceedings of the 58th Annual Meeting of the Association
  for Computational Linguistics, {ACL} 2020, Online, July 5-10, 2020}, pages
  7871--7880. Association for Computational Linguistics.

\bibitem[{Liu et~al.(2022)Liu, Shen, Zhang, Dolan, Carin, and Chen}]{kategpt}
Jiachang Liu, Dinghan Shen, Yizhe Zhang, Bill Dolan, Lawrence Carin, and Weizhu
  Chen. 2022.
\newblock \href {https://doi.org/10.18653/v1/2022.deelio-1.10} {What makes good
  in-context examples for gpt-3?}
\newblock In \emph{Proceedings of Deep Learning Inside Out: The 3rd Workshop on
  Knowledge Extraction and Integration for Deep Learning Architectures,
  DeeLIO@ACL 2022, Dublin, Ireland and Online, May 27, 2022}, pages 100--114.
  Association for Computational Linguistics.

\bibitem[{Lu et~al.(2022)Lu, Bartolo, Moore, Riedel, and
  Stenetorp}]{lu-etal-2022-fantastically}
Yao Lu, Max Bartolo, Alastair Moore, Sebastian Riedel, and Pontus Stenetorp.
  2022.
\newblock \href {https://doi.org/10.18653/v1/2022.acl-long.556} {Fantastically
  ordered prompts and where to find them: Overcoming few-shot prompt order
  sensitivity}.
\newblock In \emph{Proceedings of the 60th Annual Meeting of the Association
  for Computational Linguistics (Volume 1: Long Papers)}, pages 8086--8098,
  Dublin, Ireland. Association for Computational Linguistics.

\bibitem[{Martin et~al.(2022)Martin, Fan, de~la Clergerie, Bordes, and
  Sagot}]{martin-etal-2022-muss}
Louis Martin, Angela Fan, {\'E}ric de~la Clergerie, Antoine Bordes, and
  Beno{\^\i}t Sagot. 2022.
\newblock \href {https://aclanthology.org/2022.lrec-1.176} {{MUSS}:
  Multilingual unsupervised sentence simplification by mining paraphrases}.
\newblock In \emph{Proceedings of the Thirteenth Language Resources and
  Evaluation Conference}, pages 1651--1664, Marseille, France. European
  Language Resources Association.

\bibitem[{Nassar et~al.(2019)Nassar, Ananda{-}Rajah, and Haffari}]{NassarAH19}
Islam Nassar, Michelle Ananda{-}Rajah, and Gholamreza Haffari. 2019.
\newblock \href {https://aclanthology.org/U19-1023/} {Neural versus non-neural
  text simplification: {A} case study}.
\newblock In \emph{Proceedings of the The 17th Annual Workshop of the
  Australasian Language Technology Association, {ALTA} 2019, Sydney, Australia,
  December 4-6, 2019}, pages 172--177. Australasian Language Technology
  Association.

\bibitem[{Papineni et~al.(2002)Papineni, Roukos, Ward, and
  Zhu}]{papineni-etal-2002-bleu}
Kishore Papineni, Salim Roukos, Todd Ward, and Wei-Jing Zhu. 2002.
\newblock \href {https://doi.org/10.3115/1073083.1073135} {{B}leu: a method for
  automatic evaluation of machine translation}.
\newblock In \emph{Proceedings of the 40th Annual Meeting of the Association
  for Computational Linguistics}, pages 311--318, Philadelphia, Pennsylvania,
  USA. Association for Computational Linguistics.

\bibitem[{Qiang et~al.(2020)Qiang, Li, Yi, Yuan, and Wu}]{qiang2020BERTLS}
Jipeng Qiang, Yun Li, Zhu Yi, Yunhao Yuan, and Xindong Wu. 2020.
\newblock Lexical simplification with pretrained encoders.
\newblock \emph{Thirty-Fourth AAAI Conference on Artificial Intelligence}, page
  8649–8656.

\bibitem[{Rello et~al.(2013)Rello, Baeza-Yates, Bott, and
  Saggion}]{10.1145/2461121.2461126}
Luz Rello, Ricardo Baeza-Yates, Stefan Bott, and Horacio Saggion. 2013.
\newblock \href {https://doi.org/10.1145/2461121.2461126} {Simplify or help?
  text simplification strategies for people with dyslexia}.
\newblock In \emph{Proceedings of the 10th International Cross-Disciplinary
  Conference on Web Accessibility}, W4A '13, New York, NY, USA. Association for
  Computing Machinery.

\bibitem[{Rubin et~al.(2022)Rubin, Herzig, and
  Berant}]{rubin-etal-2022-learning}
Ohad Rubin, Jonathan Herzig, and Jonathan Berant. 2022.
\newblock \href {https://doi.org/10.18653/v1/2022.naacl-main.191} {Learning to
  retrieve prompts for in-context learning}.
\newblock In \emph{Proceedings of the 2022 Conference of the North American
  Chapter of the Association for Computational Linguistics: Human Language
  Technologies}, pages 2655--2671, Seattle, United States. Association for
  Computational Linguistics.

\bibitem[{Sheang and Saggion(2021)}]{sheang-saggion-2021-controllable}
Kim~Cheng Sheang and Horacio Saggion. 2021.
\newblock \href {https://aclanthology.org/2021.inlg-1.38} {Controllable
  sentence simplification with a unified text-to-text transfer transformer}.
\newblock In \emph{Proceedings of the 14th International Conference on Natural
  Language Generation}, pages 341--352, Aberdeen, Scotland, UK. Association for
  Computational Linguistics.

\bibitem[{Sorensen et~al.(2022)Sorensen, Robinson, Rytting, Shaw, Rogers,
  Delorey, Khalil, Fulda, and Wingate}]{SorensenRRSRDKF22}
Taylor Sorensen, Joshua Robinson, Christopher~Michael Rytting, Alexander~Glenn
  Shaw, Kyle~Jeffrey Rogers, Alexia~Pauline Delorey, Mahmoud Khalil, Nancy
  Fulda, and David Wingate. 2022.
\newblock \href {https://doi.org/10.18653/v1/2022.acl-long.60} {An
  information-theoretic approach to prompt engineering without ground truth
  labels}.
\newblock In \emph{Proceedings of the 60th Annual Meeting of the Association
  for Computational Linguistics (Volume 1: Long Papers), {ACL} 2022, Dublin,
  Ireland, May 22-27, 2022}, pages 819--862. Association for Computational
  Linguistics.

\bibitem[{Stajner(2021)}]{stajner-2021-automatic}
Sanja Stajner. 2021.
\newblock \href {https://doi.org/10.18653/v1/2021.findings-acl.233} {Automatic
  text simplification for social good: Progress and challenges}.
\newblock In \emph{Findings of the Association for Computational Linguistics:
  ACL-IJCNLP 2021}, pages 2637--2652, Online. Association for Computational
  Linguistics.

\bibitem[{Sulem et~al.(2018)Sulem, Abend, and Rappoport}]{sulem-etal-2018-bleu}
Elior Sulem, Omri Abend, and Ari Rappoport. 2018.
\newblock \href {https://doi.org/10.18653/v1/D18-1081} {{BLEU} is not suitable
  for the evaluation of text simplification}.
\newblock In \emph{Proceedings of the 2018 Conference on Empirical Methods in
  Natural Language Processing}, pages 738--744, Brussels, Belgium. Association
  for Computational Linguistics.

\bibitem[{Tanprasert and Kauchak(2021)}]{tanprasert-kauchak-2021-flesch}
Teerapaun Tanprasert and David Kauchak. 2021.
\newblock \href {https://doi.org/10.18653/v1/2021.gem-1.1} {Flesch-kincaid is
  not a text simplification evaluation metric}.
\newblock In \emph{Proceedings of the 1st Workshop on Natural Language
  Generation, Evaluation, and Metrics (GEM 2021)}, pages 1--14, Online.
  Association for Computational Linguistics.

\bibitem[{Williams et~al.(2003)Williams, Reiter, and
  Osman}]{williams-etal-2003-experiments}
Sandra Williams, Ehud Reiter, and Liesl Osman. 2003.
\newblock \href {https://aclanthology.org/W03-2317} {Experiments with
  discourse-level choices and readability}.
\newblock In \emph{Proceedings of the 9th {E}uropean Workshop on Natural
  Language Generation ({ENLG}-2003) at {EACL} 2003}, Budapest, Hungary.
  Association for Computational Linguistics.

\bibitem[{Xu et~al.(2016)Xu, Napoles, Pavlick, Chen, and
  Callison-Burch}]{xu-etal-2016-optimizing}
Wei Xu, Courtney Napoles, Ellie Pavlick, Quanze Chen, and Chris Callison-Burch.
  2016.
\newblock \href {https://doi.org/10.1162/tacl_a_00107} {Optimizing statistical
  machine translation for text simplification}.
\newblock \emph{Transactions of the Association for Computational Linguistics},
  4:401--415.

\bibitem[{Zhang et~al.(2020)Zhang, Kishore, Wu, Weinberger, and
  Artzi}]{bertscore}
Tianyi Zhang, Varsha Kishore, Felix Wu, Kilian~Q. Weinberger, and Yoav Artzi.
  2020.
\newblock \href {https://openreview.net/forum?id=SkeHuCVFDr} {Bertscore:
  Evaluating text generation with {BERT}}.
\newblock In \emph{8th International Conference on Learning Representations,
  {ICLR} 2020, Addis Ababa, Ethiopia, April 26-30, 2020}. OpenReview.net.

\bibitem[{Zhang et~al.(2022)Zhang, Feng, and Tan}]{ZhangFT22}
Yiming Zhang, Shi Feng, and Chenhao Tan. 2022.
\newblock \href {https://aclanthology.org/2022.emnlp-main.622} {Active example
  selection for in-context learning}.
\newblock In \emph{Proceedings of the 2022 Conference on Empirical Methods in
  Natural Language Processing, {EMNLP} 2022, Abu Dhabi, United Arab Emirates,
  December 7-11, 2022}, pages 9134--9148. Association for Computational
  Linguistics.

\bibitem[{Zhao et~al.(2021)Zhao, Wallace, Feng, Klein, and
  Singh}]{zhao2021calibrate}
Zihao Zhao, Eric Wallace, Shi Feng, Dan Klein, and Sameer Singh. 2021.
\newblock \href {http://proceedings.mlr.press/v139/zhao21c.html} {Calibrate
  before use: Improving few-shot performance of language models}.
\newblock In \emph{Proceedings of the 38th International Conference on Machine
  Learning, {ICML} 2021, 18-24 July 2021, Virtual Event}, volume 139 of
  \emph{Proceedings of Machine Learning Research}, pages 12697--12706. {PMLR}.

\end{thebibliography}
\bibliographystyle{acl_natbib}

\clearpage
\appendix

\section{Optimal Settings}
\label{sec:optimalsettings}

In this section, we provide full optimal settings in Table~\ref{tab:bestsettings} for the ``Random Best'' and ``MBL Best'' models. These settings are all in-domain (i.e., ASSET Validation, ASSET Tune; MTurk Validation, MTurk Tune) and include model size, $k$ (number of examples), and ordering (high/low, low/high, and random).

\begin{table}[!htp]
\scalebox{0.65}{
\begin{tabular}{l|l|l|l|l} \toprule
Metric and Dataset  & Model & Metric   & k  & Ordering \\ \midrule
MBL Best TurkCorpus   & GPT-175B & SARI & 6  & High/Low \\
Random Best TurkCorpus & GPT-175B & SARI & 8  & Low/High \\
MBL Best ASSET        & GPT-175B & SARI & 15 & Random   \\
Random Best ASSET      & GPT-175B & SARI & 10 & Random  \\
\bottomrule
\end{tabular}
}
\caption{MBL-Best and Random-Best settings for results}
\label{tab:bestsettings}
\end{table}

\section{BLEU Results}
\label{sec:bleuresults}

\begin{figure*}[ht]
    \centering  
    \includegraphics[scale=0.45]{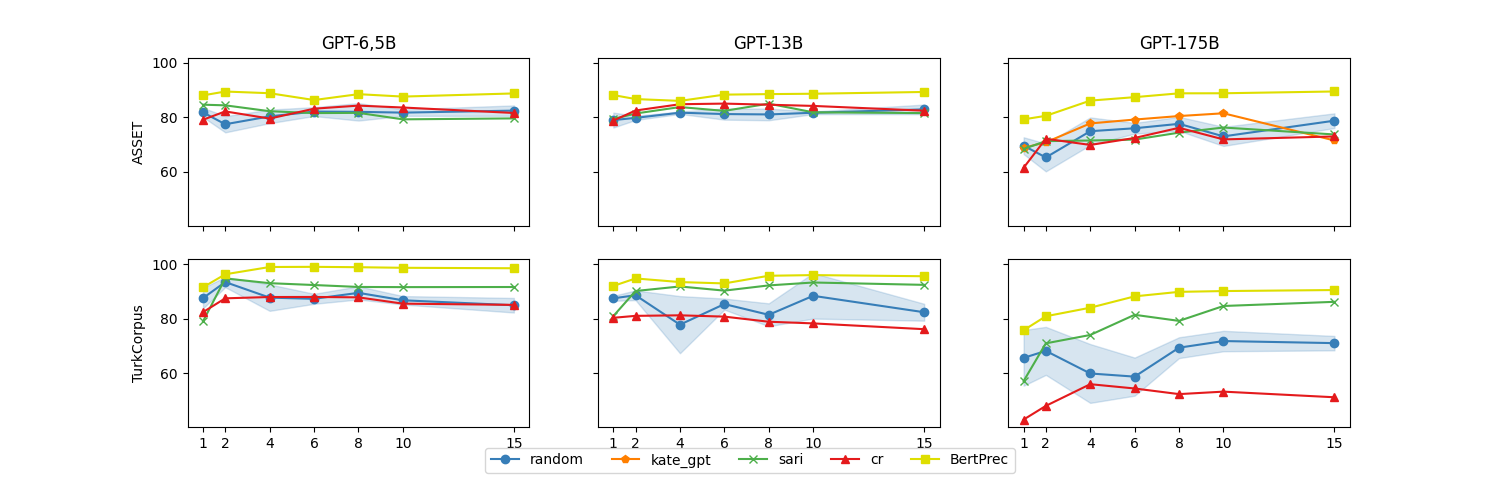}
    \caption{BLEU scores for GPT-6.5, GPT-13B and GPT-175B models on ASSET (top) and TurkCorpus (bottom) datasets.}
    \label{fig:dataset_all_models_bleu}
\end{figure*}

In this section, we include full BLEU results in Figure \ref{fig:dataset_all_models_bleu} including all model sizes ($175$B, $13$B, $6.7$B), datasets (TurkCorpus and ASSET) and selection techniques. 

\section{Top ASSET Examples}
\label{sec:topasset}

In this section, we include extended results from \ref{tab:top2examples}, with the top $3$-$10$ results from the ASSET Validation dataset based on top SARI (Section \ref{ssec:top10sari}), CR, and BERTPrec (Section \ref{ssec:top10bertprec}) scores.

\subsection{SARI}
\label{ssec:top10sari}

In this section, we include the top $3$ to $10$ examples based on SARI score selected from the ASSET Validation dataset in Table~\ref{tab:3to10sari}.

\begin{table*}[htbp]
\centering
\small
\renewcommand{\arraystretch}{1.5}
\begin{tabular}{p{0.05\linewidth} p{0.75\linewidth}} \toprule
k & Example \\ \midrule
3 & \textbf{Complex Sentence} It is adjacent to Lord Wandsworth College. \newline \textbf{Simple Sentence} it is next to Lord Wandsworth College.\\
\midrule
4 & \textbf{Complex Sentence} He took the post of chief conductor of the Netherlands Radio Philharmonic in 1957. \newline \textbf{Simple Sentence} He became the chief conductor of the Netherlands Radio Philharmonic in 1957.\\
\midrule
5 & \textbf{Complex Sentence} It was discovered on February 27, 1995. \newline \textbf{Simple Sentence} It was found on February 27, 1995.\\
\midrule
6 & \textbf{Complex Sentence} Surnames Aaron Schock, member of the U. S. House of Representatives representing the 18th district of Illinois. \newline \textbf{Simple Sentence} Surnames Aaron Schock is a member of the U. S. House of Representatives. He represents the 18th district of Illinois.\\
\midrule
7 & \textbf{Complex Sentence} Mork holds a Professorship at the Norwegian Academy of Music, Oslo. \newline \textbf{Simple Sentence} Mork is a Professor at the Norwegian Academy of Music.\\
\midrule
    8   & \textbf{Complex Sentence} The Hubble Space Telescope observed Fortuna in 1993.  \newline                                   \textbf{Simple Sentence}  The Hubble Space Telescope saw Fortuna in 1993.\\
    \midrule

    9   & \textbf{Complex Sentence} The lithosphere is underlain by the asthenosphere, the weaker, hotter, and deeper part of the upper mantle.  \newline                                   \textbf{Simple Sentence} The lithosphere is supported by the asthenosphere, the weaker, hotter, and deeper part of the upper mantle.\\
    \midrule

    10   & \textbf{Complex Sentence} The Beatles famously included his face on the cover of Sgt. Pepper's Lonely Hearts Club Band (Guy and Llewelyn-Jones 2004, 111).
  \newline                                   \textbf{Simple Sentence} The Beatles put his face on the cover of Sgt. Pepper's Lonely Hearts Club Band.\\ 
    \bottomrule
    \end{tabular}
    \caption{Top 3-10 Examples from SARI, ASSET Validation dataset.} 
    \label{tab:3to10sari}
\end{table*}

\subsection{BERTPrec}
\label{ssec:top10bertprec}

In this section, we include the top $3$ to $10$ examples based on SARI score selected from the ASSET Validation dataset in Table~\ref{tab:bertprecexamples}.

\begin{table*}[htbp]
\centering
\small
\renewcommand{\arraystretch}{1.5}
\begin{tabular}{p{0.05\linewidth} p{0.75\linewidth}} \toprule
k & Example \\ \midrule
3 & \textbf{Complex Sentence} The Convent has been the official residence of the Governor of Gibraltar since 1728. \newline \textbf{Simple Sentence} The Convent has been the residence of the Governor of Gibraltar since 1728.\\
\midrule
4 & \textbf{Complex Sentence} Scholarships, Academic Awards, Flying Eagle Awards and Improvement Awards are given to students with outstanding academic achievements. \newline \textbf{Simple Sentence} Scholarships, Academic Awards, Flying Eagle Awards and Improvement Awards are given to students with academic achievements.\\
\midrule
5 & \textbf{Complex Sentence} The blood vessels in the human body include arteries, veins and capillaries. \newline \textbf{Simple Sentence} The blood vessels in the human body are called arteries, veins and capillaries.\\
\midrule
6 & \textbf{Complex Sentence} Frederick had a summer residence built there for Sophie Charlotte by the architect Johann Arnold Nering between 1695 and 1699. \newline \textbf{Simple Sentence} Frederick had a summer residence built for Sophie Charlotte by the architect Johann Arnold Nering between 1695 and 1699.\\
\midrule
7 & \textbf{Complex Sentence} The film stars Al Pacino, John Cazale, Chris Sarandon, James Broderick, and Charles Durning. \newline \textbf{Simple Sentence} The movie stars Al Pacino, John Cazale, Chris Sarandon, James Broderick, and Charles Durning.\\
\midrule
8 & \textbf{Complex Sentence} According to an interview in the UK newspaper The Sun, Heyman wrote the brand's weekly scripts and submitted them to writers for possible changes, and then Vince McMahon for final approval. \newline \textbf{Simple Sentence} According to an interview in the UK newspaper The Sun, Heyman wrote the brand's weekly scripts and sent them to writers for possible changes, and then Vince McMahon for final approval.\\
\midrule
9 & \textbf{Complex Sentence} In March 2001, the World Wrestling Federation purchased World Championship Wrestling. \newline \textbf{Simple Sentence} In March 2001, the World Wrestling Federation bought World Championship Wrestling.\\
\midrule
10 & \textbf{Complex Sentence} Becker defeated Jim Courier in straight sets to win the 1992 year-end ATP Tour World Championships in Frankfurt. \newline \textbf{Simple Sentence} Becker defeated Jim Courier in straight sets to win the 1992 year-end ATP Tour World Championships.\\
\bottomrule
\end{tabular}
\caption{Top 3-10 Examples from BERTPrec, ASSET Validation dataset.}
\label{tab:bertprecexamples}
\end{table*}

\section{Selected Model Generated Outputs}
\label{sec:modeloutputs}

In this section, we analyze select model generated outputs on 4 (5 for GPT-175B) different example-selection methods (Random, SARI, CR, BERTPrec and optionally KATE-GPT) on different models for in-domain configurations of the ASSET dataset. The original sentence in all of these is "OEL manga series Graystripe's Trilogy There is a three volume original English-language manga series following Graystripe, between the time that he was taken by Twolegs in Dawn until he returned to ThunderClan in The Sight." This sentence was specifically picked from the ASSET/TurkCorpus test dataset based on three reasons: 1) complexity (potentially confusing abbreviations and unconventional sentence structure) 2) length 3) unfamiliar/domain-specific terms from "Warrior Cats" (e.g. "ThunderClan" and "Twolegs"). \S\ref{ssec:curieasset} includes generations on GPT-$13$B, \S\ref{ssec:babbageasset} includes generatons on GPT-$6.7$B, and \S\ref{ssec:dv3asset} includes generations on GPT-$175$B.

\subsection{Selected GPT-13B Generations}
\label{ssec:curieasset}
In this section, we include generations from the original sentence mentioned in \ref{sec:modeloutputs} on GPT-$13$B on the ASSET Test set.

\begin{table*}[htbp]
         \centering
            \small
             \renewcommand{\arraystretch}{1.5}
        \begin{tabular}{p{0.1\linewidth} p{0.75\linewidth}} \toprule
        Metric   & Curie Simplifications \\   \midrule                                                                                                     
        Random   &  there is a three volume manga series following graystripe between the time that he was taken by twolegs in dawn and he returned to thunderclan in the sight .\\
        \midrule
        SARI   &  There is a three-volume original English-language manga series following graystripe, between the time he was taken by twolegs in dawn and he returned to thunderclan in the sight.\\
        \midrule
        CR   &  There is a three volume manga series following graystripe between the time he was taken by twolegs in dawn and he returned to thunderclan in the sight.\\
        \midrule
        BERTPrec   &  There is a three volume original English-language manga series following graystripe , between the time that he was taken by twolegs in dawn and he returned to thunderclan in the sight .\\ 
        \bottomrule
        \end{tabular}
        \caption{Curie Simplifications for a selected complex sentence, ran on ASSET.} 
         \label{tab:curie}
    \end{table*}

\subsection{Selected GPT-6.7B Generations}
\label{ssec:babbageasset}
In this section, we include generations from the original sentence mentioned in \ref{sec:modeloutputs} on GPT-$6.7$B on the ASSET Test set.

\begin{table*}[htbp]
            \small  
             \renewcommand{\arraystretch}{1.5}
        \begin{tabular}{p{0.1\linewidth} p{0.75\linewidth}} \toprule
        Metric   & Curie Simplifications \\   \midrule                                                                                                     
        Random   &  oel manga series graystripe 's trilogy there is a three volume original english-language manga series following graystripe between the time that he was taken by twolegs in dawn until he returned to thunderclan in the sight .\\
        \midrule
        SARI   &  oel manga series graystripe 's trilogy there is a three volume original english-language manga series following graystripe , between the time that he was taken by twolegs in dawn until he returned to thunderclan in the sight .\\
        \midrule
        CR   &  OEL manga series, Graystripe 's Trilogy, has a three volume original English-language manga series following Graystripe, between the time that he was taken by twolegs in dawn until he returned to thunderclan in the sight.
\\
        \midrule
        BERTPrec   &  oel manga series graystripe 's trilogy there is a three volume original english-language manga series following graystripe , between the time that he was taken by twolegs in dawn until he returned to thunderclan in the sight .\\ 
        \bottomrule
        \end{tabular}
        \caption{Babbage Simplifications for a selected complex sentence, ran on ASSET.} 
        \label{tab:babbage}
    \end{table*}

\subsection{Selected GPT-175B Generations}
\label{ssec:dv3asset}
In this section, we include generations from the original sentence mentioned in \ref{sec:modeloutputs} on GPT-$175$B on the ASSET Test set. Text in \textcolor{Mycolor3}{red} indicates text that has been successfully been changed from the abbreviation "OEL" to an interpretable phrase (either "English-language" or "original English-language"). 

\begin{table*}[t]
\centering \small
\renewcommand{\arraystretch}{2}
\begin{tabular}{@{}lp{0.80\textwidth}@{}}
\toprule
{\bf Original}    & OEL manga series Graystripe's Trilogy There is a three volume original English-language manga series following Graystripe, between the time that he was taken by Twolegs in Dawn until he returned to ThunderClan in The Sight.\\
{\bf Reference Simplification}      & There is a three volume \textcolor{Mycolor3}{original English-language} manga following Graystripe, between the time that he was taken by Twolegs in Dawn until he returned to ThunderClan in The Sight.\\
\hline
 \\
\midrule
{\bf k=2}    &  \\
{\bf KATE-GPT}      & Graystripe's Trilogy is a three volume \textcolor{Mycolor3}{original English-language} manga series following Graystripe from when he was taken by Twolegs to his return to ThunderClan.\\
\hline
{\bf BERTPrec}      & Graystripe's trilogy is a three-volume \textcolor{Mycolor3}{English-language} manga series about his time between being taken by twolegs in Dawn and returning to ThunderClan in The Sight.\\
\hline
{\bf SARI}      & Graystripe's trilogy is a three-volume \textcolor{Mycolor3}{original English-language} manga series. It follows Graystripe between the time he was taken by Twolegs in Dawn and when he returned to ThunderClan in The Sight.
\\
\hline
 \\
\midrule
{\bf k=8}    &  \\
{\bf KATE-GPT}      & The Graystripe's Trilogy is an OEL manga series that follows Graystripe's journey between the time he was taken by Twolegs in Dawn and his return to ThunderClan in The Sight.
\\
\hline
{\bf BERTPrec}      & Oel manga series Graystripe's trilogy follows Graystripe between being taken by twolegs in Dawn and returning to ThunderClan in The Sight. \\
\hline
{\bf SARI}      & Graystripe's trilogy is a three volume \textcolor{Mycolor3}{English-language} manga series about his time away from ThunderClan in Dawn. \\
\hline
 \\
\midrule
{\bf k=15}    &  \\
{\bf KATE-GPT}      & The OEL manga series Graystripe's Trilogy follows Graystripe between the time he was taken by Twolegs in Dawn and when he returned to ThunderClan in The Sight.\\
\hline
{\bf BERTPrec}      & Oel manga series graystripe's trilogy is a three-volume English-language manga series following Graystripe between his capture by twolegs in Dawn and his return to ThunderClan in The Sight.\\
\hline
{\bf SARI}      & Graystripe's trilogy is a three-volume \textcolor{Mycolor3}{English-language} manga series. It covers the time Graystripe was taken by twolegs in Dawn until he returned to ThunderClan in The Sight.
\\
\bottomrule
\end{tabular}
\caption{Model-generated simplifications on ASSET, GPT-$175$B.}
\label{tab:examples}
\end{table*}

\end{document}